\documentclass[letterpaper, 10 pt, conference]{ieeeconf}  

\IEEEoverridecommandlockouts                              

\usepackage{amsmath,amssymb,amsfonts}
\usepackage{multirow}
\usepackage{graphicx}
\usepackage{rotating}
\usepackage{multicol}
\usepackage{amsmath}
\usepackage{float}
\usepackage{array}
\usepackage{svg}
\usepackage{caption}
\usepackage{subcaption}
\usepackage{siunitx}
\usepackage{float}
\usepackage{xcolor}
\usepackage{multirow}
\usepackage{graphicx}
\usepackage{color, soul, colortbl}
\usepackage{nicematrix}
\usepackage{colortbl} 
\usepackage{url}
\usepackage{hyperref}

\DeclareUnicodeCharacter{2061}{}

\usepackage{fancyhdr}
\fancypagestyle{withfooter}{
  
  \fancyfoot[C]{\footnotesize Accepted to the IEEE IROS workshop on Autonomous Robotic Systems in Aquaculture: Research Challenges and Industry Needs}
}

\title{\LARGE \bf
Autonomous Underwater Robotic System for Aquaculture Applications
}

\author{Waseem Akram, Muhayyuddin Ahmed, Lakmal Seneviratne, and Irfan Hussain$^{*}$                                                     
\thanks{ Khalifa University Center for Autonomous Robotic Systems (KUCARS), Khalifa University, United Arab Emirates.}%
\thanks{$^{*}$ Corresponding Author, Email: irfan.hussain@ku.ac.ae}
}

\begin{document}

\maketitle
\thispagestyle{withfooter}
\pagestyle{withfooter}

\begin{abstract}
Aquaculture is a thriving food-producing sector producing over half of the global fish consumption. However, these aquafarms pose significant challenges such as biofouling, vegetation, and holes within their net pens and have a profound effect on the efficiency and sustainability of fish production. Currently, divers and/or remotely operated vehicles are deployed for inspecting and maintaining aquafarms; this approach is expensive and requires highly skilled human operators. This work aims to develop a robotic-based automatic net defect detection system for aquaculture net pens oriented to on-ROV processing and real-time detection of different aqua-net defects such as biofouling, vegetation, net holes, and plastic. The proposed system integrates both deep learning-based methods for aqua-net defect detection and feedback control law for the vehicle movement around the aqua-net to obtain a clear sequence of net images and inspect the status of the net via performing the inspection tasks. This work contributes to the area of aquaculture inspection, marine robotics, and deep learning aiming to reduce cost, improve quality, and ease of operation.

\end{abstract}


\section{Introduction}

According to the World Economic Forum, nearly 90 percent of the world’s wild fish stocks are now fully exploited, overexploited or depleted (\url{https://datatopics.worldbank.org/}). Fish is an important source of a wide range of essential micronutrients and is a crucial part the daily diets of billions of people around the world. At present, fish consumption per capita in the United Arab Emirates (UAE) stands at 22.5 kg, whereas aquaculture provides merely 2\% to the total annual fish consumption (\url{https://www.foodsecurity.gov.ae/}). Consequently, there is a growing need to establish large-scale aquaculture farms in the open sea of the UAE, aimed to mitigate the supply shortage. Aquaculture is a central component of the UAE’s National Food Security Strategy and heavy investments are being made in developing aquaculture projects and infrastructure. 

\begin{figure}[t]
\centering
\includegraphics[width=1\linewidth]{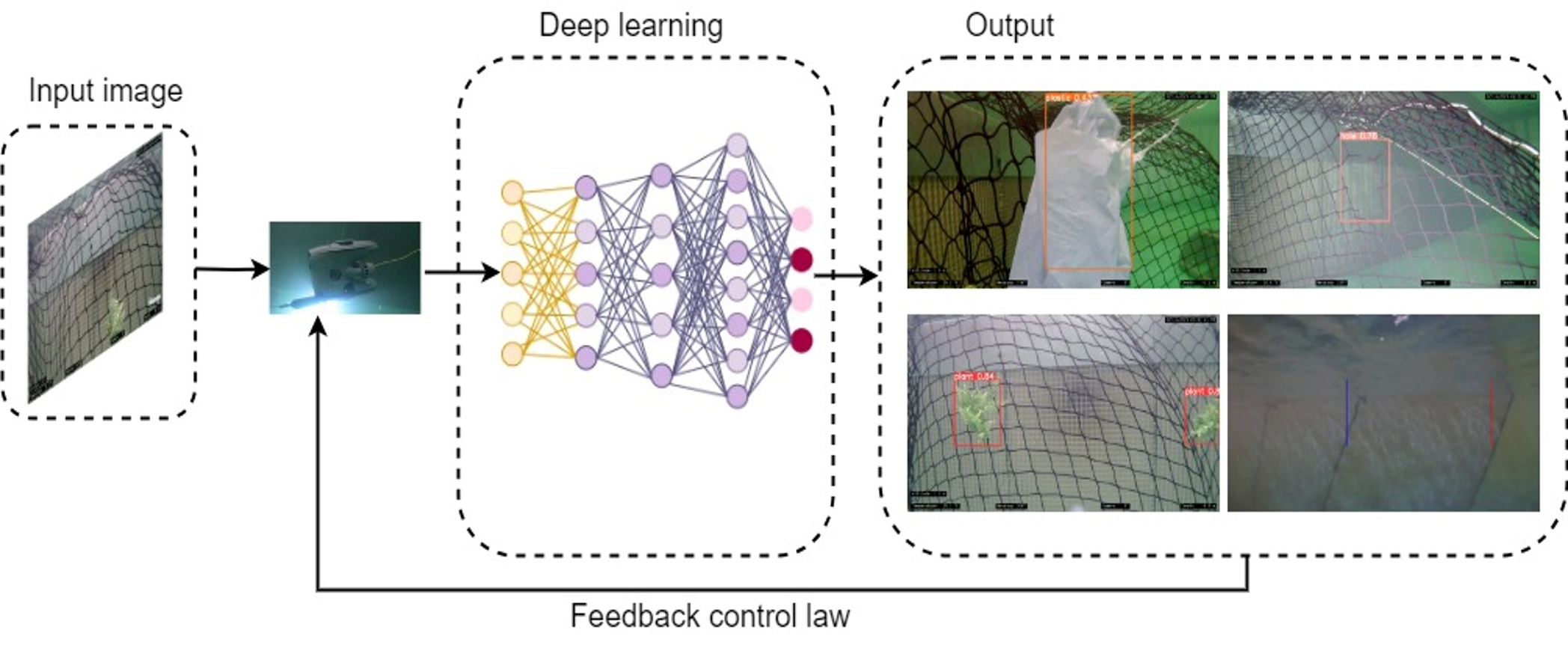}
\caption{Aquaculture net pens inspection project concept. An ROV takes camera input, apply deep learning model, perform aqua-net defect detection, and gets reference input via feedback control law for tracking the net plane.}
\label{fig:block}
\end{figure}

\begin{figure*}[t]
\centering
\includegraphics[width=1\linewidth]{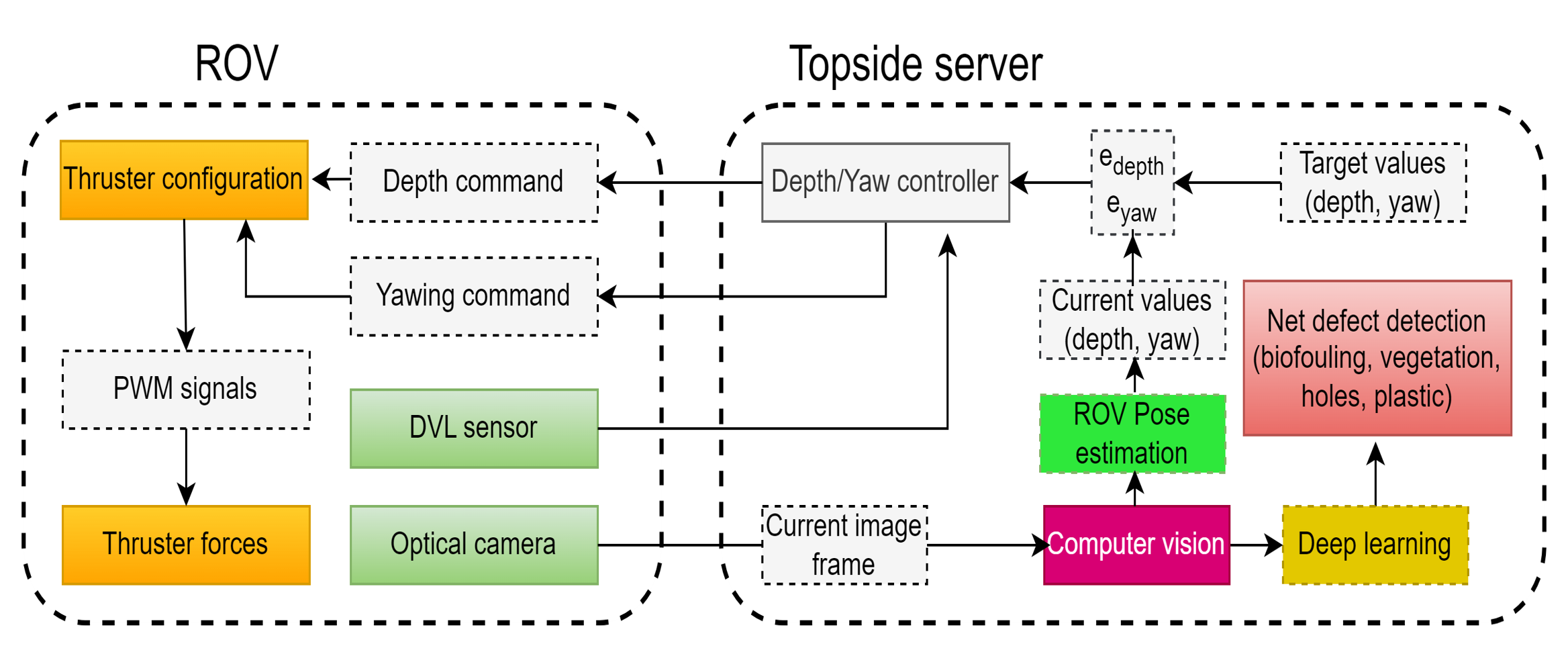}
\caption{Proposed block diagram of the system. 
}
\label{fig:method}
\end{figure*}

Marine aquaculture has become an essential part of meeting the growing demand for high-quality protein while also preserving the ocean environment \cite{gui2019research}. There are various types of marine aquaculture facilities, such as deep-sea cage farming, raft farming, deep-sea platform farming, and net enclosure farming \cite{yan2018research}. However, the netting used in these facilities is a crucial component that is easily damaged and difficult to detect, leading to the escape of cultured fish and causing substantial economic losses every year \cite{zhou2018current}. Also, in marine environments, several living organisms like bacteria, algae, and hybdroids attach themselves to the enclosures, i.e., the bio-fouling phenomenon, which is detrimental to the fish and the structural integrity of the enclosures \cite{comas2018fouling}. The detection of net damage is currently impeding the development of marine aquaculture facility \cite{wei2020intelligent,bakht2023mula}.

Traditionally, inspection and maintenance of aquaculture involve the deployment of divers or the use of video feeds obtained through Remotely Operated Vehicles (ROVs). However, this method is both costly and poses significant risks. Furthermore, it provides low coverage, verifiability, and repeatability \cite{akram2022visual}. In contrast, analyzing images captured by ROVs equipped with different types of cameras eliminates safety risks and offers the potential to implement computer-assisted damage detection procedures through computer vision techniques. In this line, the application of deep learning approaches includes Mask R-RCNN \cite{maskrcnn}, Fast-RCNN \cite{fastrcnn}, YOLO \cite{yolov5}, and SSD \cite{ssd} for automating aquaculture net damage detection  \cite{sun2020deep,yang2021deep,albitar2016underwater,akram2024aquaculture}. For instance, Zhang et al. \cite{zhang2022netting} proposed the use of Mask-RCNN \cite{maskrcnn} for net hole detection problems in a laboratory setup, achieving an impressive Average Precision score of 94.48\%. Likewise, Liao et al. \cite{liao2022research} proposed a MobileNet-SSD network model for hole detection in open-sea fish cages. They achieved a commendable average precision score of 88.5\% by combining MobileNet with the SSD network model for net hole detection. Tao et al. \cite{tao2018omnidirectional}, applied deep learning to detect net holes using the YOLOv1 (You Only Look Once) algorithm on images captured under controlled lab conditions. In contrast, Madshaven et al. \cite{madshaven2022hole}, employed a combination of classical computer vision and image processing techniques for tracking, alongside neural networks for segmenting the net structure and classifying scene content, including the detection of irregularities caused by fish or seaweed. Furthermore, Qiu et al. \cite{qiu2020fishing}, utilized image-enhancing methods for net structure analysis and marine growth segmentation in their work.

The existing aquaculture inspection solutions using ROV depend on teleoperation and most of the studies have focused on the hole detection problem. Teleoperation of ROV-based inspections requires highly skilled teleoperators to control the position and trajectory of the underwater vehicle. ROV movements due to waves and currents act as unwanted disturbances and degrade the overall system performance in collecting high-quality data for inspection tasks \cite{akram2021visual}. Therefore, it is important to improve the autonomy of existing robotic systems to reduce costs and improve accuracy and ease of use, so that they can be routinely utilized for the inspection (e.g., fouling detection, vegetation detection, and plastic detection) of aquafarms. Moreover, net inspection-related approaches either use datasets recorded at fisheries for benchmarking algorithms or deploy actual vehicle(s) at net pens. Thus, most of them are regarded as working in unstructured environments

\subsection{Contributions}

In this study, we present a novel underwater robotic approach aimed at automating net inspection tasks in aquaculture as shown in Figure \ref{fig:block}. The proposed method combines deep learning with the ROV to efficiently detect irregularities in fish-nets during aqua-net inspections.

Specifically, the approach utilizes a deep learning-based detector to identify areas within the net structure that may contain potential defects like plants, holes, or plastic. Additionally, a closed-loop control law guides the ROV along the net plane, from top to bottom, using traditional computer vision techniques such as edge detection. This enables the detection of reference points based on camera images to determine the vehicle's position relative to the net plane, and allow the ROV navigation around the net plane.

By implementing this proposed system, the inspection process becomes autonomous, significantly reducing the need for human divers or operators. This not only enhances the safety of the production process but also directly safeguards the health of laborers. As a result, the aquaculture industry can increase its added value in fish farming while ensuring the protection of its workforce.

\section{Proposed method}

This work strives to equip the local aquaculture industry by implementing cutting-edge inspection and monitoring technology. This involves utilizing an ROV in conjunction with deep learning techniques. Figure \ref{fig:method} illustrates the concept behind this initiative. The system comprises two main components: the Blueye Pro ROV X platform and a topside server unit.

To establish communication and control, the ROV connects to the topside server on the surface through WiFi. Once deployed into the sea, the ROV operates autonomously, executing commands received from the topside server.

The automatic inspection system integrates vision-based navigation and net defect detection systems. First, the ROV navigates autonomously around the fish net sending video sequences to the topside server. The ROV navigation is achieved by employing a computer vision algorithm on the input image obtained from the forward-looking ROV’s camera.  From the topside server unit, the velocity commands are sent to the ROV, which controls the ROV’s yaw and depth. The control module is responsible to control the ROV movement and allow traversing of the fishnet pen with a target distance in real-time. For the ROV position estimation, we make use of two parallel ropes attached to the net surface as a region of interest. By computer vision method, we perform the edge detection, estimate the ROV relative distance to the net, and develop a motion controller that generates the thruster inputs, which make the ROV follow the net pen resulting in a complete net inspection task. On the topside server unit, the detection module determines the net status by using the deep learning-based detection method for the identification of different net defects such as biofouling, vegetation, holes, and plastic debris. The entire inspection process follows to the subsequent steps.

\begin{itemize}
    \item At each time step, based on the camera input, estimate the ROV distance to the net plane.
    \item  Control the ROV distance and yaw angle relative to the net plane.
    \item Perform net defect detection tasks by applying a deep learning-based detector method.
    \item Through vision servoing and velocity control, traverse the net surface from top to bottom with a target distance and speed.
\end{itemize}

\subsection{ROV distance approximation and Control}
\begin{figure}[t]
\centering
\includegraphics[width=1\linewidth]{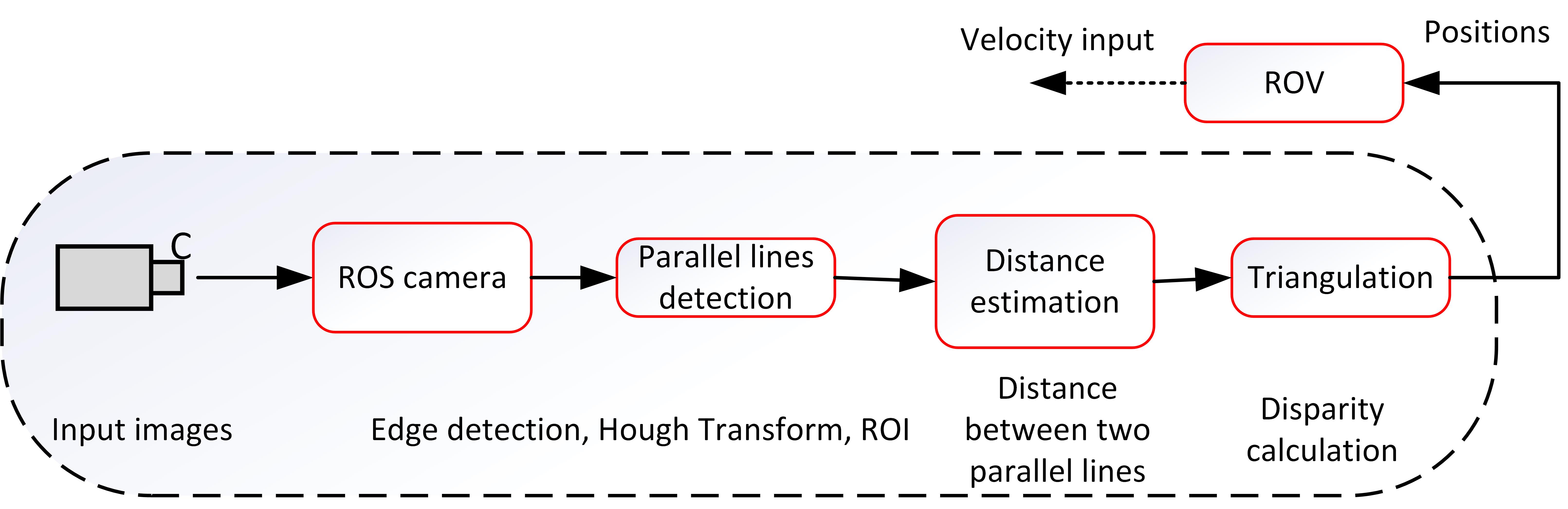}
\caption{ROV distance approximation and control.}
\label{fig:control}
\end{figure}

The control module  as shown in Figure \ref{fig:control} comprises a reference path generation and a control law to govern the vehicle's movement. The reference path is established by identifying two parallel ropes attached to the net in the input image. Subsequently, the position of the ropes in the image is estimated to approximate the ROV's distance relative to the net plane. The ropes can be considered as parallel straight lines, and their detection is accomplished using the edge detection Canny and Hough transform method. Using the computed pixel distance and knowing the real distance between the ropes, positions are determined through triangulation method \cite{akram2022visual}. These positions serve as essential inputs for vehicle control and navigation algorithm.

\subsection{Net defect detection approach}

\begin{figure}[t]
\centering
\subfloat[Biofouling.  \label{fig:B2}]{\includegraphics[width=0.5\columnwidth, height=0.3\columnwidth]{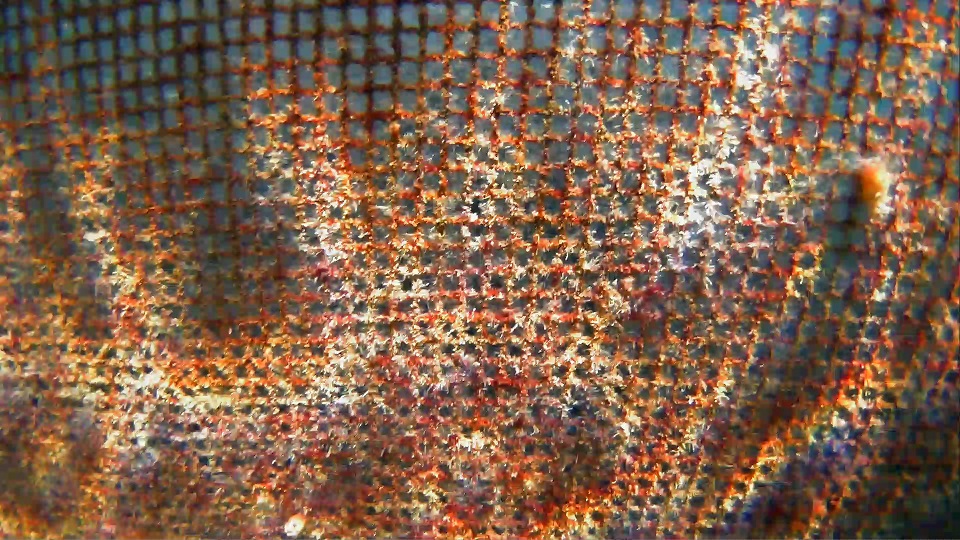}}
\subfloat[Net hole. \label{fig:B1}]{\includegraphics[width=0.5\columnwidth,height=0.3\columnwidth]{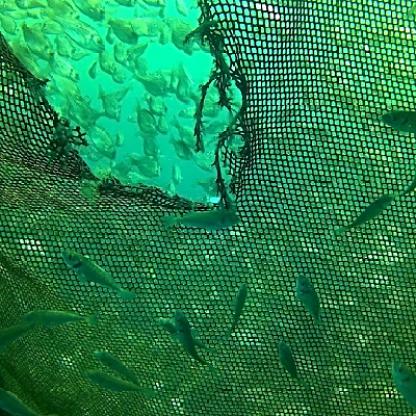}}
\vspace{5pt}\hspace{0.2cm}
\subfloat[Plastic. \label{fig:B3}]{\includegraphics[width=0.5\columnwidth, height=0.3\columnwidth]{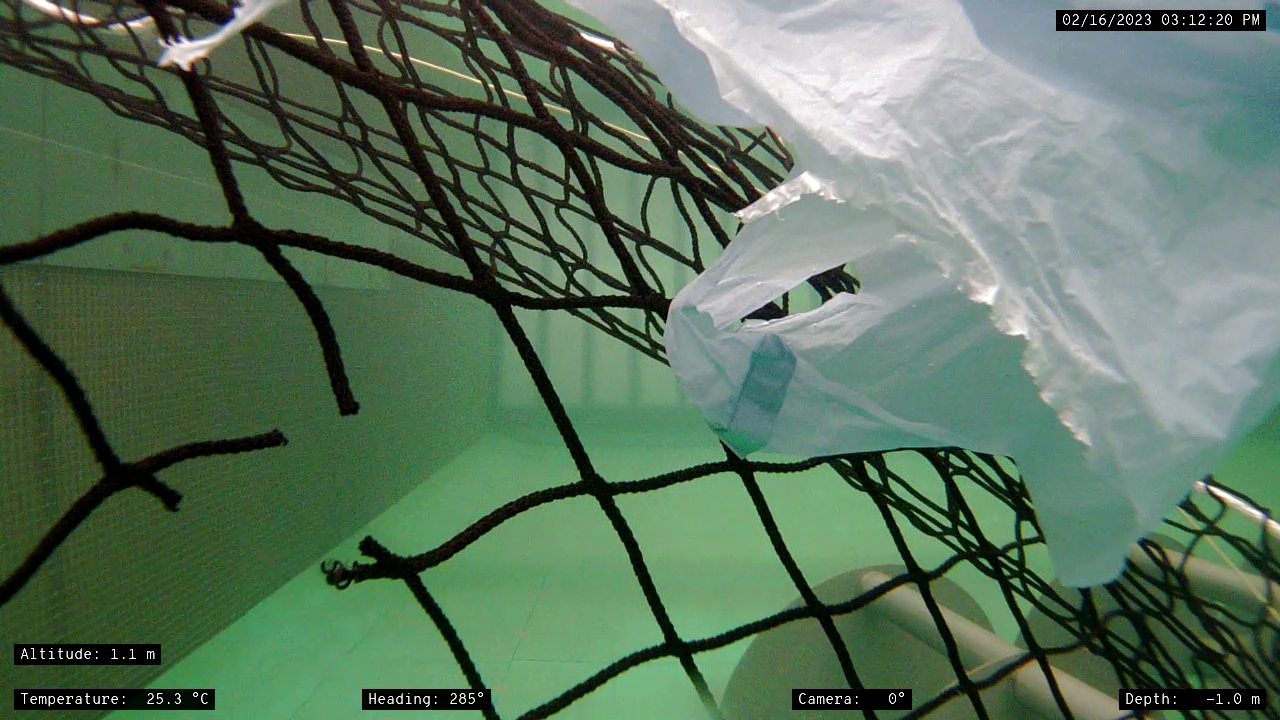}}
\subfloat[Vegetation.  \label{fig:B4}]{\includegraphics[width=0.5\columnwidth, height=0.3\columnwidth]{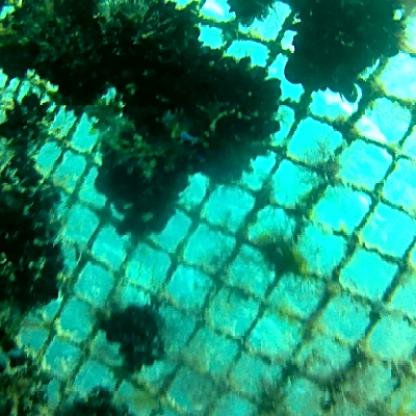}}\vspace{5pt}\hspace{0.2cm}
\caption{Aquaculture net pens defect dataset.}
\label{fig:dataset}
\end{figure}

Deep learning-based methods are highly effective in recognizing and locating the defects in the imagery data. In this work, we adopted a deep learning model named YOLOV5 \cite{yolov5} for identification and localization of vegetation, plastic, and holes in the aquaculture net pens. As an example, some instances of the dataset are shown in Figure \ref{fig:dataset}.

\begin{figure}[t]
\centering
\includegraphics[width=1\linewidth]{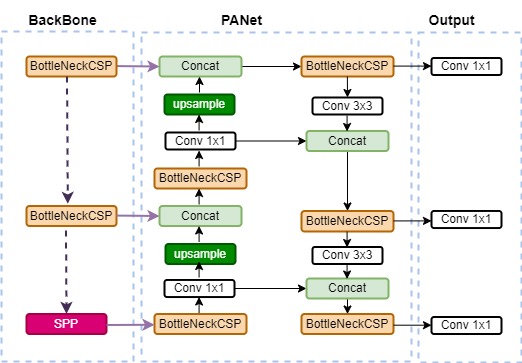}
\caption{The architectural diagram of YOLOv5 \cite{yolov5}.}
\label{fig:yolo}
\end{figure}

The employed network model YOLOV5 (Figure \ref{fig:yolo}) consists of a backbone network responsible to extract features from the input image. These extracted features are then given to the other several layers, including convolutional, umpsampling, and fusion layers, that generate high-resolution feature maps. The model further utilizes anchor boxes, which are pre-defined boxes of various sizes and aspect ratios, to predict bounding boxes for objects. YOLOv5 predicts the coordinates of bounding boxes relative to the grid cells and refines them with respect to anchor boxes. To train the YOLOv5 model, a large labeled dataset is required, along with bounding box annotations for the objects of interest. The model is trained using techniques like backpropagation and gradient descent to optimize the network parameters and improve object detection performance.

\section{Results and Discussions}

\begin{figure}[t]
\centering
\includegraphics[width=1\linewidth]{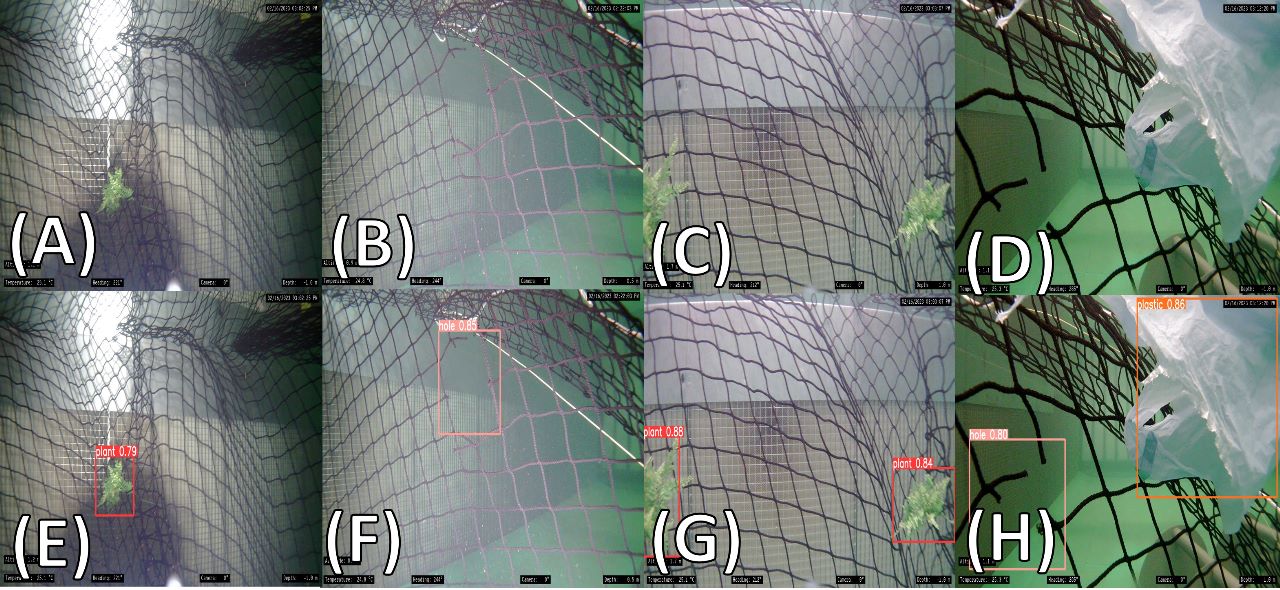}
\caption{Results obtained during the inspection campaign of aquaculture net defect detection.}
\label{fig:res-detect}
\end{figure}

\begin{figure}[t]
\centering
\includegraphics[width=1\linewidth]{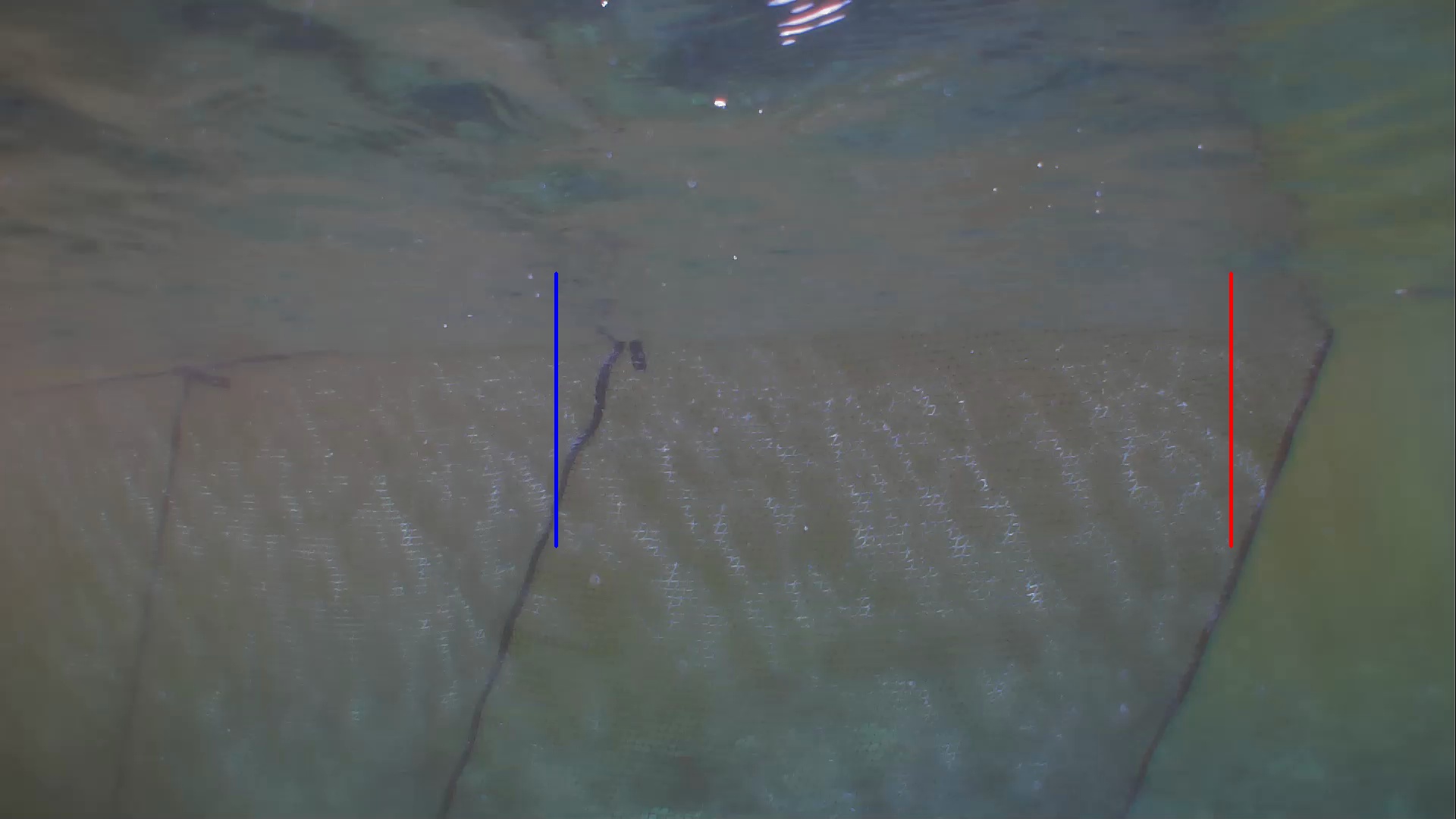}
\includegraphics[width=1\linewidth]{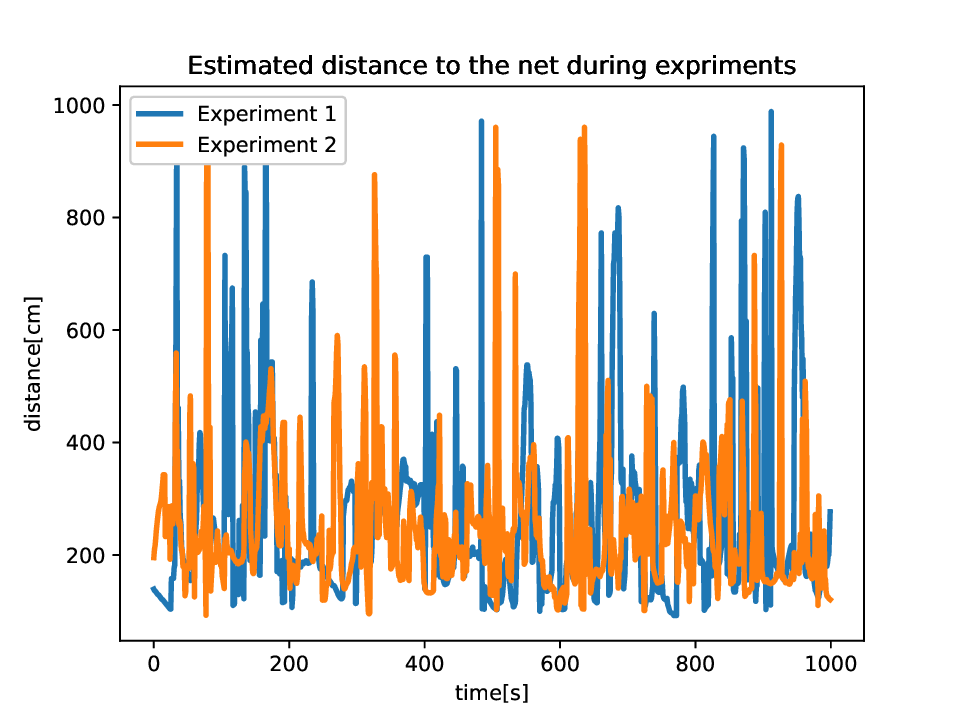}
\caption{Distance of the ROV relative to the aqua-net plane during the experiments. Part (a) shows the detection of reference point and Part (b) shows the estimated ROV distance.}
\label{fig:res-depth}
\end{figure}

In this section, we discuss the implementation results. The aqua-net defect visualization results are shown in Figure \ref{fig:res-detect}. Here, the analysis was performed to view the detected aqua-net defects i.e., plant, holes, and plastic after training the YOLOv5 model. The model was deployed on the unseen real-time images in a pool set up to check the feasibility of the model after training on the collected custom dataset. Moreover, the model showed the ability to detect the net defect of different sizes as can be noted in Figure \ref{fig:res-detect}, where the first row (A-D) shows the input images and the second row (E-H) shows the corresponding results. Subsequently, the outcomes of the control module are demonstrated in Figure \ref{fig:res-depth}. Evidently, the algorithm effectively recognizes the ropes as a pair of parallel lines within the input. This recognition serves as a reference point for distance estimation and guides the vehicle's path during navigation. The experimental results, showed in Figure \ref{fig:res-depth} (b), illustrate the estimated distances. Throughout the experiments, a reference distance of 200 cm was employed. The vehicle's forward/backward movements were then dictated based on the disparity between the current distance and the reference, facilitating the execution of the inspection task \cite{din2023marine,ahmed2023vision}.

\section{Conclusion}\label{sec:conclusion}

This paper introduces ROV coupled deep learning method that focuses on detecting irregularities in fish nets, which is a critical task within the full inspection process. The method proposed in this paper employs a deep learning-based detector to identify areas where potential net defects such as plants, holes, and plastic exist in the net structure. Moreover, the system also incorporates a feedback control law by making use of vision based approach for vehicle movement around the net plane. The method has been tested at a lab pool setup for fish nets. The experimental results demonstrated the applicability of the proposed solution for aquaculture net inspection activities. 

\section*{Acknowledgement}
\noindent This publication is based upon work supported by the Khalifa University of Science and Technology under Award No. CIRA-2021-085, FSU-2021-019, RC1-2018-KUCARS.

\bibliographystyle{ieeetr}
\bibliography{refs}

\end{document}